\documentclass{article}
\usepackage[preprint]{neurips_2024} 

\usepackage{amsmath, amssymb, amsfonts}
\usepackage{graphicx}

\usepackage{algorithm}
\usepackage{algpseudocode}

\usepackage[utf8]{inputenc} 
\usepackage[T1]{fontenc}    
\usepackage{hyperref}       
\usepackage{url}            
\usepackage{booktabs}       
\usepackage{amsfonts}       
\usepackage{nicefrac}       
\usepackage{microtype}      
\usepackage{xcolor}         

\title{PARIS: Pruning Algorithm via the Representer theorem for Imbalanced Scenarios}

\author{
  Enrico Camporeale \\
  Department of Physics and Astronomy \\
  Queen Mary University of London \\
  London, United Kingdom \\
  and\\
  SWx-TREC \\
  (Space Weather Technology Research and Education Center) \\
  University of Colorado, Boulder, CO, USA \\
  \texttt{enrico.camporeale@qmul.ac.uk} \\
}

\begin{document}

\maketitle

\begin{abstract}
The challenge of \textbf{imbalanced regression} arises when standard Empirical Risk Minimization (ERM) biases models toward high-frequency regions of the data distribution, causing severe degradation on rare but high-impact ``tail'' events. Existing strategies uch as loss re-weighting or synthetic over-sampling often introduce noise, distort the underlying distribution, or add substantial algorithmic complexity.

We introduce \textbf{PARIS} (Pruning Algorithm via the Representer theorem for Imbalanced Scenarios), a principled framework that mitigates imbalance by \emph{optimizing the training set itself}. PARIS leverages the representer theorem for neural networks to compute a \textbf{closed-form representer deletion residual}, which quantifies the exact change in validation loss caused by removing a single training point \emph{without retraining}. Combined with an efficient Cholesky rank-one downdating scheme, PARIS performs fast, iterative pruning that eliminates uninformative or performance-degrading samples. 

We use a real-world space weather example, where PARIS reduces the training set by up to 75\% while preserving or improving overall RMSE, outperforming re-weighting, synthetic oversampling, and boosting baselines. Our results demonstrate that representer-guided dataset pruning is a powerful, interpretable, and computationally efficient approach to rare-event regression.
\end{abstract}

\section{Introduction}

In many high-stakes domains — from environmental sciences and financial forecasting to healthcare — training data is inherently imbalanced and follows long-tailed distributions. 
While modern deep learning thrives on vast datasets, the standard principle of Empirical Risk Minimization (ERM) tends to bias models toward the majority regions of the data manifold, often leading to poor generalization on rare but critical ``tail'' events. 
Addressing this \emph{imbalanced regression} problem typically relies on heuristic strategies such as re-weighting loss functions \citep{ren2018learning, yang2021delving, steininger2021density} or re-sampling techniques like SMOGN \citep{branco2017smogn, torgo2013smote}. 
However, these methods can introduce noise, lead to overfitting on minority samples, or require complex hyperparameter tuning. 
A more fundamental question remains: rather than artificially inflating the data or modulating the loss, can we identify and retain only the minimal set of specific training examples that structurally support generalization across the entire target distribution?

Understanding how individual training samples contribute to model behavior is essential for answering this question. 
Recent advances in sample attribution, particularly \emph{influence functions} \citep{koh2017understanding} and \emph{representer point methods} \citep{yeh2018representer}, provide principled mechanisms to decompose a model’s prediction into a linear combination of its training data. 
In this paper, we introduce \textbf{PARIS} (Pruning Algorithm via the Representer theorem for Imbalanced Scenarios), a unified framework that leverages the representer theorem to iteratively refine training data for imbalanced regression. 
By analytically quantifying the ``structural value'' of each sample via a validation set, our approach constructs specialized data subsets that excel at rare-event prediction without requiring synthetic data generation. 
The proposed method builds conceptually upon three complementary lines of research: training data attribution, dataset pruning, and the representer theorem for neural networks.

\subsection{Training data attribution and influence analysis}

The classical framework of \emph{influence functions}, rooted in robust statistics \citep{hampel1986robust, cook1982residuals}, was adapted to deep learning by \citet{koh2017understanding} to approximate the effect of perturbing or removing a single training point. 
Formally, this relies on the inverse Hessian of the loss, allowing to identify which training points are most responsible for a specific prediction. 
While conceptually powerful, Hessian-based influence methods suffer from instability and prohibitive computational costs in modern deep networks \citep{basu2021influence}. 
Recent work has attempted to mitigate these costs through approximate unrolling of gradient descent trajectories \citep{pruthi2020estimating, bae2024training} or higher-order jackknife techniques \citep{alaa2020discriminative}. 
However, theoretical analyses suggest that influence estimates can be brittle in large-scale non-convex settings \citep{schioppa2023theoretical, epifano2023revisiting}, and standardized benchmarking reveals significant variance in their reliability \citep{bareeva2024quanda, hammoudeh2024training}. 
For imbalanced regression, where minority samples may have high leverage but also high variance, stable and tractable attribution is crucial.

\paragraph{The Representer Theorem for neural networks.}

To overcome the computational bottlenecks of Hessian inversion, a parallel line of research exploits the \emph{representer theorem}. 
Originating in kernel methods \citep{scholkopf2001representer, shawe2004kernel}, the theorem establishes that for a broad class of regularized optimization problems, the optimal solution lies in the span of the training data evaluations. 
\citet{yeh2018representer} extended this insight to Deep Neural Networks (DNNs) by treating the penultimate layer as a learned feature map $\phi(x)$. 
For a network with an $\ell_2$-regularized linear final layer, the prediction for a test point $x_t$ can be decomposed exactly as:
\begin{equation}
    f(x_t) = \sum_{j=1}^{N} \alpha_j \, k(x_j, x_t) = \sum_{j=1}^{N} \alpha_j \, \phi(x_j)^\top \phi(x_t),
\end{equation}
where $\phi(x_j)$ denotes the learned feature embedding of training sample $j$, and $\alpha_j$ is a scalar representer coefficient derived from the training loss. 
Unlike influence functions, this decomposition is exact given the fixed features $\phi$, and the coefficients $\alpha_j$ explicitly quantify the contribution of training sample $x_j$ to the prediction $f(x_t)$. 
This formulation provides a closed-form, differentiable link between training data and model output. 
Crucially, it allows for the efficient estimation of \emph{leave-one-out} residuals—estimating how the loss would change if a sample were removed—without costly retraining, forming the backbone of our proposed pruning algorithm.

\paragraph{Dataset Pruning.}

Our work utilizes these attribution scores to perform \emph{dataset pruning}, the practice of removing training samples to improve efficiency or generalization. 
Pruning methods typically fall into three categories: geometry-based, uncertainty-based, and error-based.
Geometry-based methods, such as Coresets \citep{feldman2010coresets, lucic2018training}, select a subset of points that cover the feature space, often disregarding the target labels. 
Uncertainty-based methods retain samples near the decision boundary (high entropy) while discarding ``easy'' examples \citep{lewis1994sequential}. 
Conversely, error-based methods \citep{ren2018learning} often discard``hard'' samples with high training loss, assuming them to be noise or outliers.

However, standard pruning heuristics struggle in the context of imbalanced regression. 
In these settings, high-error samples are often the rare, informative targets we wish to learn, not noise. 
Blindly keeping high-error samples can lead to overfitting, while keeping only low-error samples exacerbates the bias toward the majority mean.
\citet{yang2022dataset} proposed pruning based on generalization influence, yet few methods explicitly optimize for distributional imbalance.
Our approach differs by using the representer theorem to identify samples that structurally improve \emph{validation performance} on a held-out set, thereby naturally selecting a balance of representative majority samples and crucial minority exemplars.

\subsection{Contributions}

In this work, we introduce a novel representer-guided pruning and selection strategy termed \textbf{Pruning Algorithm via the Representer theorem for Imbalanced Scenarios}. PARIS iteratively refines the training data by identifying and removing samples that are least beneficial (or most detrimental) to generalization performance, specifically targeting improvement in imbalanced regression settings.

Our formulation allows for highly efficient iterative updates by leveraging matrix identities derived from the representer theorem, similar to techniques used in kernel ridge regression \citep{rifkin2007notes}. Crucially, this avoids the high computational cost of full model retraining or recomputation of representer terms during the pruning cycle. The proposed approach exploits the analytic structure of the representer theorem to compute, in closed form, the exact change in validation residuals that would result from removing any single training sample.

The paper provides the following key methodological and empirical contributions:

\begin{itemize}
    \item \textbf{Closed-Form Representer Deletion Residual:} We derive a novel closed-form expression for the representer deletion residual. This formula analytically quantifies the impact ($\Delta \mathcal{L}$) of removing a single training point on the validation set predictions and the model's overall loss \textbf{without requiring model retraining}. This enables fast, principled evaluation of sample importance, forming the core of the PARIS selection criterion.

    \item \textbf{Greedy Iterative Pruning Algorithm:} We introduce a \textbf{greedy iterative pruning algorithm} that leverages rank-one update techniques (e.g., Cholesky downdating). This allows us to sequentially remove the $K$ worst-performing samples efficiently. By integrating the deletion residual with incremental representer coefficient ($\alpha$) and feature matrix ($\Phi$) updates, the cost of removing $K$ points is reduced from $O(K \cdot N^3)$ (for standard matrix inversion) to a near-linear cost in $N$ per step, where $N$ is the current dataset size.

    \item \textbf{Empirical Validation on Space Weather Forecasting:} We demonstrate the effectiveness of PARIS on a real-world, highly imbalanced regression benchmark: forecasting the {geomagnetic $D_{st}$ index}

We show that PARIS achieves strong data efficiency, significantly reducing the required training dataset size (down to 25\% of the original size) while maintaining or improving predictive accuracy in the most critical tail events (extreme geomagnetic storms) compared to state-of-the-art baselines.
\end{itemize}

\section{Methodology}

\subsection{Representer formulation for neural networks}
For completeness, we derive here the representer formulation for neural networks, following closely the formalism of \citep{yeh2018representer}.
We decompose the neural network into a feature extractor  $\phi(x) \in \mathbb{R}^{D}$ (all layers up to the penultimate layer) and a linear output layer with weights $w \in \mathbb{R}^{D}$. The prediction for an input $x$ is
\begin{equation}
\hat{y}(x) = \phi(x)^{\top} w ,
\end{equation}
where any bias term is absorbed into $\phi(x)$.
Keeping the features $\phi(x)$ fixed, this can then be interpreted as a linear problem.
Given training data $\{(x_j,y_j)\}_{j=1}^{N}$, the final layer is trained via $\ell_2$-regularized least squares:
\begin{equation}
\mathcal{L}(w)
= \frac{1}{2} \sum_{j=1}^{N} \left( \phi(x_j)^{\top} w - y_j \right)^{2}
+ \frac{\lambda}{2} \| w \|^{2},
\end{equation}
with optimal solution
\begin{equation}
\label{eq:wstar}
w^* = \left( \Phi^{\top}\Phi + \lambda I_D \right)^{-1} \Phi^{\top} Y ,
\end{equation}
where $\Phi \in \mathbb{R}^{N\times D}$ contains the feature vectors as rows. It is important to emphasize that the weights $w_{NN}$ learned by training the NN via back-propagation are not necessarily equal to $w^*$. However, one can estimate the value of the regularization parameter $\lambda$ that minimizes their discrepancy (see Appendix~A). We assume, and empirically demonstrate below, that such small discrepancy does not ultimately affect the validity of the PARIS approach. 

Using the matrix inversion lemma,
\[
(\Phi^{\top}\Phi + \lambda I_D)^{-1}\Phi^{\top}
= \Phi^{\top}(\Phi\Phi^{\top} + \lambda I_N)^{-1},
\]
the prediction for a test point $x_t$ becomes
\begin{equation}
\hat{y}_t
= \phi_t^{\top} \Phi^{\top} (\Phi\Phi^{\top} + \lambda I_N)^{-1} Y .
\end{equation}
This motivates the definition of the dual coefficients
\begin{equation}
\alpha = (\Phi\Phi^{\top} + \lambda I_N)^{-1} Y \in \mathbb{R}^{N}.
\end{equation}

Let $\tilde{\phi}(x_j) = \alpha_j \phi(x_j)$ and $\tilde{\Phi} = \Phi \odot \alpha$ be the matrix containing these scaled features. Then the prediction assumes the explicit representer form
\begin{equation}
\label{eq:representer-new}
\boxed{
\hat{y}_t = \sum_{j=1}^{N} \phi(x_t)^{\top} \tilde{\phi}(x_j)
}
\end{equation}
which decomposes each test prediction into the contributions of individual training samples. 
Effectively,  $\tilde{\Phi}$ (the matrix whose $j$-th row is $\tilde{\phi}(x_j)$) represents the basis function learned by the neural network. The representation in Eq. \ref{eq:representer-new} is extremely powerful and transparent in quantifying the contribution of each training point $x_j$ to the neural network output $\hat{y}_t$. For instance, it is obvious that any point $x_j$ for which the vectors $\tilde{\phi}(x_j)$ and $\phi(x_t)$ are orthogonal, does not contribute to $\hat{y}_t$.

For a test set $\{x_{t_i}\}_{i=1}^{N_{\text{test}}}$ with feature matrix $\Phi_t$, we define the \textbf{Representer Influence Matrix}
\begin{equation}
\label{eq:S-matrix-new}
S = \Phi_t \tilde{\Phi}^{\top} \in \mathbb{R}^{N_{\text{test}} \times N_{\text{train}}},
\end{equation}
where $S_{i,j}$ is the contribution of training point $x_j$ to the prediction at $x_{t_i}$. 
The final prediction for $x_{t_i}$ is the sum of the contributions along the $i$-th row of $\mathbf{S}$:
\begin{equation}
\hat{y}_{t_i} = \sum_{j=1}^N S_{i, j}
\end{equation}
The matrix $S$ forms the foundation for our pruning metric, as removing a training point $x_k$ is equivalent to setting the $k$-th column of ${S}$ to zero.

\subsection{Pruning via representer influence}

The representer form \eqref{eq:representer-new} and influence matrix \eqref{eq:S-matrix-new}
provide a transparent mechanism for pruning. 
The definition of $S$ allows for the implementation of various pruning strategies by analyzing its rows and columns. For instance, removing redundant points corresponds to finding training points $x_k$ that contribute minimally across all test points (small magnitude ${S}_{i, k}$ for all $i$), while improving accuracy on challenging samples requires focusing on points that strongly influence prediction errors. In this paper, we focus explicitly on improving the performance of imbalanced regression tasks, hence our pruning strategy will be guided by iteratively reducing the residual $r_{v^*}$ of the validation point $x_{v^*}$ that exhibits the largest current squared residual loss, identified by the index $v^*$:
\[
r_{v^*} = y_{v^*} - \hat{y}_{v^*}.
\]
Because $\hat{y}_{v^*} = \sum_{j=1}^{N} S_{v^*,j}$, removing a training point $x_k$ is equivalent to zeroing the $k$-th column of $S$. The updated prediction and residual are
\begin{align}
\hat{y}_{v^*}^{\setminus k} &= \hat{y}_{v^*} - S_{v^*,k}, \\
r_{v^*}^{\setminus k} &= y_{v^*} - \hat{y}_{v^*}^{\setminus k}
= r_{v^*} + S_{v^*,k},
\end{align}
where the symbol $\setminus k$ denotes the point $x_k$ being pruned.
By defining the loss associated with a single point as $\mathcal{L}_{v^*} = r_{v^*}^2$, the exact change in validation loss is
\begin{equation}
\label{eq:delta-loss-final}
\boxed{
\Delta \mathcal{L}_{v^*}^{\setminus k}
= (r_{v^*} + S_{v^*,k})^{2} - r_{v^*}^{2}
= 2 r_{v^*} S_{v^*,k} + S_{v^*,k}^{2}.
}
\end{equation}
A \emph{negative} $\Delta \mathcal{L}_{v^*}^{\setminus k}$ indicates that $x_k$ is detrimental and should be removed.

\subsubsection{Efficient updates via Cholesky rank-one downdating}
Executing the pruning process iteratively requires efficiently recalculating the dual coefficients $\alpha$ after each point removal. A full re-solve of the dual system $\alpha$ would incur $\mathcal{O}(N^{3})$ time per step and is computationally prohibitive due to the large size $N$.

Instead, we work in the primal space. Define the Gram matrix
\[
A = \Phi^{\top}\Phi + \lambda I_D, \qquad A = LL^{\top},
\]
where $L$ is its Cholesky factor. The optimal weights satisfy $w^* = A^{-1}\Phi^{\top}Y$ and can be obtained via two triangular solves.

When a training point $x_k$ with feature vector $\phi_k$ is removed, the Gram matrix becomes
\[
A_{\text{new}} = A - \phi_k \phi_k^{\top},
\]
a rank-one \emph{downdate}. The updated Cholesky factor $L_{\text{new}}$ is computed in $\mathcal{O}(D^{2})$ using standard stable algorithms. The updated weights $w_{\text{new}}^*$ then follow from solving
\[
A_{\text{new}} w_{\text{new}}^* = \Phi_{\setminus k}^{\top} Y_{\setminus k}.
\]

Because the dual coefficients satisfy $\alpha = \Phi w^*$, the updated $\alpha$ is obtained with no matrix inversion. This keeps the per-step pruning cost dominated by the $\mathcal{O}(D^{2})$ downdate, enabling efficient large-scale pruning.

A central computational object in PARIS is the influence matrix
\[
S = \Phi_{\text{val}} \, (\Phi \odot \alpha)^\top \in \mathbb{R}^{N_{\text{val}} \times N_{\text{train}}},
\]
whose $j$-th column quantifies the sensitivity of validation predictions to the removal of training point $j$.  
Naively recomputing $S$ at every pruning step requires a full matrix multiplication of cost 
$\mathcal{O}(N_{\text{val}} \, D \, N_{\text{train}})$, which dominates the runtime when $D$ is large.

To avoid this, we introduce a cached matrix
\[
T \;=\; \Phi_{\text{val}} \, \Phi^\top ,
\]
which contains \emph{unscaled} feature inner products and can be computed once per outer iteration.  
Since
\[
S_{:,j} \;=\; \alpha_j \, T_{:,j},
\]
the influence matrix can be reconstructed from $T$ using a simple column–wise scaling by the dual coefficients.  
This reduces the update cost to
\[
\mathcal{O}(N_{\text{val}}\,N_{\text{train}}),
\]
saving an entire factor of $D$ per update.  
After pruning a point $k$, the corresponding column of $T$ is removed, and a new vector of dual coefficients 
$\alpha_{\text{new}}$ is computed via a Cholesky rank-one downdate in the primal system.  
The updated influence matrix then satisfies
\[
S_{\text{new}} = T_{\setminus k} \; \operatorname{diag}(\alpha_{\text{new}}),
\]
constructed in a single vectorized operation.

This strategy provides a numerically stable and highly efficient update rule: the expensive feature inner products are computed only once per retraining cycle, while each pruning step requires merely a column deletion in $T$ and a lightweight rescaling.  
The only assumption is that the feature map $\Phi$ remains fixed throughout the inner pruning loop; this assumption is enforced by the outer loop of PARIS, which periodically retrains the feature extractor to keep $\Phi$ consistent after many pruning steps.  
Together with the adaptive choice of the regularization parameter $\lambda$ (see Algorithm~\ref{alg:paris} and Appendix~A), this yields a fast and fully incremental implementation of representer-based pruning.

\subsection{PARIS: Pruning Algorithm via the Representer
theorem for Imbalanced Scenarios}

We now summarize the complete pruning procedure,  in Algorithm \ref{alg:paris}. The process is iterative, with an outer loop that retrains the feature extractor and an inner loop that performs fast, incremental pruning based on the deletion residual $\Delta \mathcal{L}_{\text{val}}$.

\begin{algorithm}
\caption{PARIS algorithm}
\label{alg:paris}
\begin{algorithmic}[1]
\Require Training set $\mathcal{D}_{\text{train}}$, Validation set $\mathcal{D}_{\text{val}}$, Pruning fraction $p$, Total pruning fraction $P_{\max}$, Iterative pruning size $K$, Regularization $\lambda$.
\State Initialize $\mathcal{D}_P \gets \mathcal{D}_{\text{train}}$, $N \gets |\mathcal{D}_{\text{train}}|$
\While{$|\mathcal{D}_P| / N > (1 - P_{\max})$} 
    \State $\mathcal{D}_{\text{current}} \gets \mathcal{D}_P$
    \State Train feature extractor $\phi$ and final layer $w^*$ on $\mathcal{D}_{\text{current}}$
    \State Extract features: $\mathbf{\Phi} \gets \phi(\mathcal{D}_{\text{current}})$, $\mathbf{\Phi}_{\text{val}} \gets \phi(\mathcal{D}_{\text{val}})$
    \State Compute Gram matrix: $\mathbf{A} \gets \mathbf{\Phi}^\top \mathbf{\Phi} + \lambda \mathbf{I}_D$
    \State Cholesky factorization: $\mathbf{A} = \mathbf{L} \mathbf{L}^\top$
    \State Solve for weights $w^*$ and dual coefficients $\alpha = \mathbf{\Phi} w^*$
    \State Build unscaled influence matrix: $\mathbf{T} \gets \mathbf{\Phi}_{\text{val}} \mathbf{\Phi}^\top$ \Comment{Size $N_\text{val} \times N_\text{train}$}
    \State Compute scaled influence matrix: $\mathbf{S} \gets \mathbf{T} \odot \alpha^\top$ \Comment{Vectorized column-wise scaling by $\alpha$}
    \State Initialize $\mathcal{K}_{\text{pruned}} \gets \emptyset$
    \For{$i = 1$ to $K = \lfloor p \cdot |\mathcal{D}_{\text{current}}| \rfloor$}
        \State Compute residuals: $\mathbf{r}_{\text{val}} \gets Y_{\text{val}} - \mathbf{S} \mathbf{1}$
        \State Identify hardest validation point: $v^* \gets \arg\max_i (\mathbf{r}_{\text{val}})_i^2$
        \State Compute deletion residuals: $\Delta \mathcal{L}_k \gets 2 r_{v^*} S_{v^*,k} + S_{v^*,k}^2$ for $k otin \mathcal{K}_{\text{pruned}}$
        \State Select sample to prune: $k^* \gets \arg\min_k \Delta \mathcal{L}_k$
        \State $\mathcal{K}_{\text{pruned}} \gets \mathcal{K}_{\text{pruned}} \cup \{k^*\}$
        \State Rank-one Cholesky downdate: $\mathbf{L} \gets \text{CholeskyDowndate}(\mathbf{L}, \phi(x_{k^*}))$
        \State Remove pruned point from features and targets: $\mathbf{\Phi} \gets \mathbf{\Phi}_{\setminus k^*}$, $Y_{\text{train}} \gets Y_{\text{train}, \setminus k^*}$
        \State Update weights: $w^* \gets \mathbf{A}_{\text{new}}^{-1} \mathbf{\Phi}^\top Y_{\text{train}}$ using $\mathbf{L}$
        \State Update dual coefficients: $\alpha \gets \mathbf{\Phi} w^*$
        \State Update scaled influence matrix: $\mathbf{S} \gets \mathbf{T}_{:, \text{remaining}} \odot \alpha^\top$ \Comment{Only scale columns corresponding to remaining points}
    \EndFor
    \State $\mathcal{D}_P \gets \mathcal{D}_P \setminus \mathcal{K}_{\text{pruned}}$
\EndWhile

\Return $\mathcal{D}_P$ \Comment{Pruned training set}
\end{algorithmic}
\end{algorithm}

A few algorithmic design choices and assumptions deserve emphasis.

\paragraph{Core algorithmic structure.}  
At the beginning of each outer cycle we fully retrain (or fine-tune) a copy of the network on the current retained training set and extract the penultimate-layer feature matrices \(\Phi_{\text{train}}\) and \(\Phi_{\text{val}}\). We then compute the dual coefficients \(\alpha\) by solving the ridge system for the current dataset (in our implementation we use a conjugate-gradient solver, \texttt{compute\_alpha\_cg}, for numerical efficiency). An inner loop iteratively identifies the validation sample \(v^\star\) with the largest squared residual, forms the single-row representer contribution \(S_{v^\star,:} = \phi_{v^\star}^\top \tilde{\Phi}\) (where \(\tilde{\Phi}\) is the row-wise scaled feature matrix), and computes the deletion residual
\[
\Delta\mathcal{L}_{\setminus k} = 2 r_{v^\star} S_{v^\star,k} + S_{v^\star,k}^2
\]
for every remaining training point \(k\). The sample minimizing \(\Delta\mathcal{L}_{\setminus k}\) is pruned; we then recompute \(\alpha\) for the reduced set and continue the inner loop until the desired per-cycle pruning budget is reached.

\paragraph{Key assumption and role of the outer loop.}  
The inner-loop pruning logic effectively assumes that the feature extractor \(\phi(\cdot)\) remains approximately constant while removing a small fraction of points (one at a time up to the per-cycle budget). This assumption is explicitly acknowledged: it is \emph{not} true in general (removing many points will change the learned representation if the network were to be retrained). The outer loop exists precisely to address this: after each pruning cycle we retrain (or fine-tune) the feature extractor on the pruned dataset so that \(\Phi\) is refreshed and representer quantities are recomputed under the updated representation. In practice this two-level design trades off correctness and cost: the inner loop performs many cheap representer-based decisions using a fixed \(\Phi\), while the outer loop corrects for the accumulated approximation error by updating \(\Phi\) and \(\alpha\).

\section{Experimental Setup and Baselines}

We perform an empirical evaluation of the PARIS algorithm by comparing its performance against state-of-the-art methods specifically designed or adapted to handle the challenges of imbalanced regression. These competing approaches typically fall into three categories: standard Empirical Risk Minimization (ERM), advanced loss re-weighting/modification, and data augmentation/re-sampling.

Our approach departs from the standard practice in the machine learning literature, where newly proposed algorithms are typically benchmarked on a small set of well-known, often low-dimensional regression datasets (the usual suspects: Boston Housing, Abalone, Wine Quality Red, and so on). Instead of relying on these relatively simple and overused benchmarks, we set a substantially higher bar for evaluation. We test PARIS on a challenging, high-dimensional, and operationally relevant real-world prediction problem from space weather forecasting \citep{camporeale2019challenge, camporeale2023artificial}.

\begin{figure}
    \centering
    \includegraphics[width=1\linewidth]{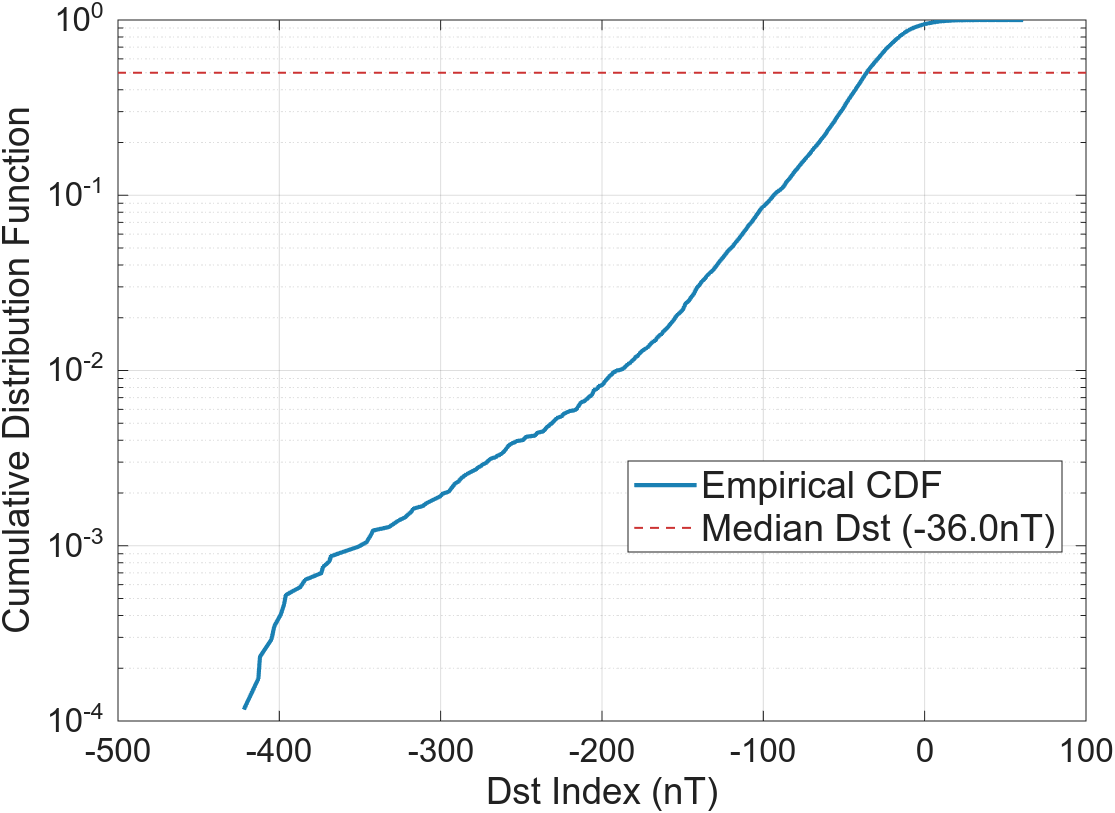}
    \caption{Cumulative Distribution Function of the Dst data used for the imbalanced regression evaluation.}
    \label{fig:dst}
\end{figure}

In particular, we evaluate the performance of PARIS on predicting the minimum Disturbance Storm Time ($D_{st}$) index, a global measure of magnetic storm severity. The $D_{st}$ index follows a severe power-law distribution, making it an archetypal imbalanced regression task where the minority samples (extreme geomagnetic storms) are the most critical to predict (see Figure \ref{fig:dst}). $D_{st}$ prediction is a long-standing and non-trivial benchmark in the space weather community (see, e.g., \cite{hu2023multi, wang2024short, wintoft2018evaluation, gruet2018multiple})

\subsubsection{Dataset and Preprocessing}
We utilize a consolidated dataset comprising $D_{st}$ index data from 100 geomagnetic storms, encompassing $17,200$ hourly data points. The input features include solar wind and interplanetary magnetic field (IMF) parameters: $B_x$, $B_y$, $B_z$, $|B|$, proton density and velocity, tilt angle, hoy (hour of the year). For each input parameter, we take a 6-hour time history, making the total input dimension equal to 56. The target is the value of $D_{st}$ one hour in the future. All data is publicly available from NASA at (\url{https://spdf.gsfc.nasa.gov/pub/data/omni/}).

The dataset is partitioned using a rigorous \textbf{Leave-One-Out Cross-Validation (LOOCV)} scheme on a subset of the 20 strongest recorded storms. In each fold, one strong storm is the test set, the 20 next strongest serve as the validation set ($\mathcal{D}_{\text{val}}$) for pruning, and the remaining storms constitute the initial training set ($\mathcal{D}_{\text{train}}$). 

\subsubsection{Baseline Models}
All models utilize the same neural network architecture (3 hidden layers: [100, 100, 50], RELU activation) and are trained using an ensemble of 20 independent runs with random seeds. Early stop is performed based on the validation loss. We compare PARIS against the following baseline models:

\begin{itemize}
    \item {\textbf{Baseline (MSE)}}
This model represents the conventional approach where the DNN is trained on the full original training dataset ($\mathcal{D}_{\text{train}}$) using the standard Mean Squared Error (MSE) loss function, $\mathcal{L}_{\text{MSE}} = \frac{1}{N} \sum_{i=1}^{N} (y_i - \hat{y}_i)^2$. 

\item{\textbf{Focal Loss (FOCAL)}}
Focal loss was originally introduced for classification to address extreme class imbalance by down-weighting easy examples and concentrating learning on hard, misclassified ones \citep{lin2017focalloss}. The idea is to modulate the base loss with a factor that increases when the model performs poorly. In regression the standard mean-squared error is computed first, and each sample’s error is normalized to the range $[0,1]$.
A focal weighting factor $\alpha \, (\mathrm{norm\_mse})^{\gamma}$ is applied to the base MSE term, where $\alpha$ controls the global loss scale and $\gamma$ modulates the emphasis on high-error (``hard'') samples.

\item{\textbf{DenseWeight Loss (DW)}}
DenseWeight estimates a density-aware importance weight for each training target by applying a kernel density estimator (KDE) in the target space and computing weights inversely proportional to the local sample density \citep{steininger2021density}. This yields larger weights for rare or under-represented target values and smaller weights for dense regions, enabling the loss function to focus more strongly on infrequent but important samples. 

\item{SMOGN}
(SMOTE for Regression with Gaussian Noise) \cite{torgo2013smote} extends the well-known SMOTE oversampling algorithm from classification to regression settings. Instead of relying on discrete class labels, SMOGN identifies ``rare'' regions in the continuous target space using a relevance function that quantifies which target values should be oversampled. New synthetic samples are then generated by interpolating between nearest neighbours in these rare regions and injecting controlled Gaussian noise to preserve variability. The intuition is that augmenting under-represented target intervals can help the model learn a more balanced error distribution and prevent bias toward the dense majority region. However, SMOGN modifies the training distribution itself via synthetic data rather than reweighting existing samples, and its effectiveness depends strongly on the relevance function, oversampling ratios, and noise parameters.

\item{XGBoost (XGB)}
Extreme Gradient Boosting (XGBoost) \citep{chen2016xgboost} is a powerful, tree-based ensemble method used as a non-neural network baseline. It is trained on the full original dataset and serves as a high-performance benchmark for structured regression.
\end{itemize}

\subsection{Results}
The Root Mean Square Error (RMSE) computed on the test set per each fold is shown in Table \ref{tab:rmse}. The last column reports the percentage of the samples pruned, with respect to the original sample size. For all experiments, we set up a hard limit of $75\%$ (that is, the iterative algorithm stops when reaching the limit) and that turned out to be the optimal pruning for a large number of folds (based on the corresponding validation set), which suggests that further pruning might have been beneficial.
The last row shows the average RMSE across folds. With the exception of SMOGN, all models present very similar performance, in terms of global RMSE. DW slightly outperforms PARIS in terms of global average RMSE. However, it is important to emphasize that the PARIS results are based on a training set that is, on average, 74\% smaller than the original training size. This clearly shows how powerful and advantageous the PARIS algorithm is.

\begin{table}
\tiny
\centering
\caption{Root Mean Square Error computed on the test set for each fold. Last column indicates the percentage of the training set pruned by PARIS. Number in bold is the best per row. Last row shows average values across folds.}\label{tab:rmse}
\begin{tabular}{lccccccc} Fold & PARIS & BASELINE & FOCAL & DW & XGB & SMOGN & Pruned (\%) \\ \toprule  01 & \textbf{29.31} & 35.65 & 38.27 & 33.76 & 56.41 & 70.18 & 75 \\  02 & \textbf{45.01} & 54.29 & 52.58 & 47.76 & 75.16 & 90.19 & 75 \\  03 & \textbf{48.63} & 52.23 & 53.44 & 50.04 & 53.08 & 62.43 & 74 \\  04 & 42.62 & 42.57 & 41.66 & \textbf{38.63} & 55.04 & 74.90 & 75 \\  05 & 20.04 & 18.12 & 20.53 & 20.99 & 23.68 & \textbf{17.74} & 74 \\  06 & \textbf{27.75} & 35.78 & 35.30 & 32.74 & 35.04 & 45.54 & 74 \\  07 & 48.31 & 41.55 & \textbf{38.36} & 41.34 & 42.28 & 49.46 & 75 \\  08 & 36.53 & 38.37 & 38.16 & \textbf{35.52} & 36.18 & 43.43 & 75 \\  09 & 22.85 & 21.56 & 24.26 & 22.04 & \textbf{21.48} & 35.04 & 72 \\  10 & 30.20 & 27.52 & \textbf{23.50} & 27.53 & 24.83 & 31.98 & 75 \\  11 & 39.41 & \textbf{28.57} & 33.40 & 29.74 & 29.26 & 44.44 & 75 \\  12 & 22.81 & 23.47 & 22.51 & \textbf{22.02} & 26.75 & 34.33 & 72 \\  13 & 23.26 & 22.34 & 21.88 & 21.89 & \textbf{20.43} & 32.47 & 72 \\  14 & 20.50 & 21.30 & 24.93 & \textbf{19.09} & 23.10 & 38.23 & 72 \\  15 & 21.55 & 20.85 & \textbf{17.74} & 18.72 & 22.57 & 35.39 & 75 \\  16 & 37.82 & 36.58 & 32.13 & 33.72 & \textbf{31.53} & 39.94 & 75 \\  17 & 25.69 & 25.93 & 24.09 & 27.52 & \textbf{16.27} & 27.53 & 75 \\  18 & \textbf{14.33} & 17.26 & 19.43 & 18.12 & 16.10 & 30.91 & 72 \\  19 & 27.08 & 23.74 & 22.41 & 23.50 & \textbf{21.58} & 25.26 & 75 \\  20 & \textbf{22.63} & 27.23 & 27.77 & 28.48 & 23.53 & 29.75 & 75 \\ \midrule \textbf{Average} & 30.32 $\pm$ 10.26 & 30.75 $\pm$ 10.88 & 30.62 $\pm$ 10.54 & \textbf{29.66 $\pm$ 9.46} & 32.71 $\pm$ 15.90 & 42.96 $\pm$ 18.33 & 74 \\ \bottomrule \end{tabular}
\end{table}

\subsubsection{Performance on Extreme Events (Conditional RMSE)}

Although used as a standard, the RMSE is not an appropriate metrics for imbalanced regression, since it is obviously dominated by non-rare samples and therefore it can be misleading in evaluating the performance of a model for the most interesting and challenging rare events. 
In this case, a more appropriate metric is the Conditional RMSE (cRMSE), which calculates the RMSE only for samples where the ground truth value is less than or equal to a threshold $T$. Table \ref{tab:cRMSE} shows the RMSE computed on given subsets of the test set, chosen as the lowest percentiles. For the most rare samples below the 20th percentile PARIS is substantially outperforming all other models.

\begin{table}
    \centering
    \caption{Conditional RMSE computed on a given percentile of the test set.}
    \label{tab:cRMSE}
\begin{tabular}{l | cccccc}Model & \multicolumn{6}{c}{RMSE on Lowest Percentiles of Dst (nT)} \\ \toprule  & \textbf{1\%} & \textbf{2\%} & \textbf{5\%} & \textbf{10\%} & \textbf{20\%} & \textbf{50\%} \\ \midrule  \textbf{PARIS} & \textbf{75.02} & \textbf{73.46} & \textbf{65.39} & \textbf{68.17} & 59.47 & 42.51 \\ \textbf{BASELINE} & 112.56 & 105.36 & 86.06 & 76.15 & 61.80 & 43.26 \\ \textbf{focal} & 114.56 & 104.57 & 84.50 & 73.38 & 57.97 & \textbf{39.22} \\ \textbf{DW} & 97.46 & 91.53 & 76.44 & 69.59 & \textbf{56.84} & 39.93 \\ \textbf{XGB} & 204.71 & 170.12 & 123.55 & 97.79 & 74.43 & 49.52 \\ \textbf{SMOGN} & 242.44 & 208.30 & 162.51 & 130.78 & 97.98 & 63.47 \\ \bottomrule \end{tabular}
\end{table}

\begin{figure}[h!]
    \centering
    \includegraphics[width=0.9\textwidth]{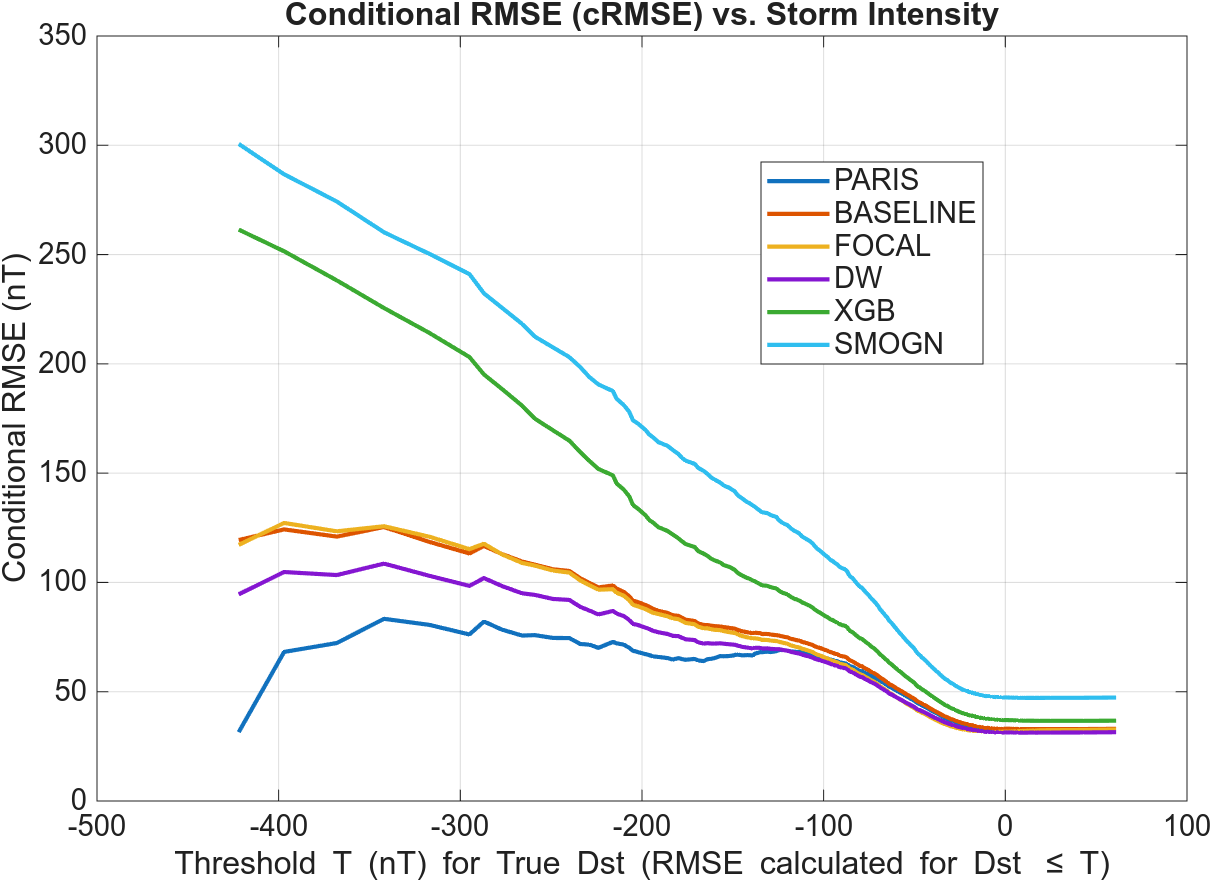}
    \caption{Conditional RMSE (cRMSE) vs. Storm Intensity Threshold $T$. cRMSE is calculated for all samples where the ground truth $D_{st} \leq T$. Lower values of $T$ (more negative $D_{st}$) represent the most severe storm events.}
    \label{fig:crmse}
\end{figure}

A graphical view of the cRMSE is presented in Figure \ref{fig:crmse}, where the threshold $T$ is in physical units (nT) rather than percentiles. The cRMSE plot clearly shows that PARIS maintains the lowest cRMSE across the most critical region ($D_{st} \leq -100$ nT).

\begin{figure}[H]
    \centering
    \includegraphics[width=0.9\linewidth]{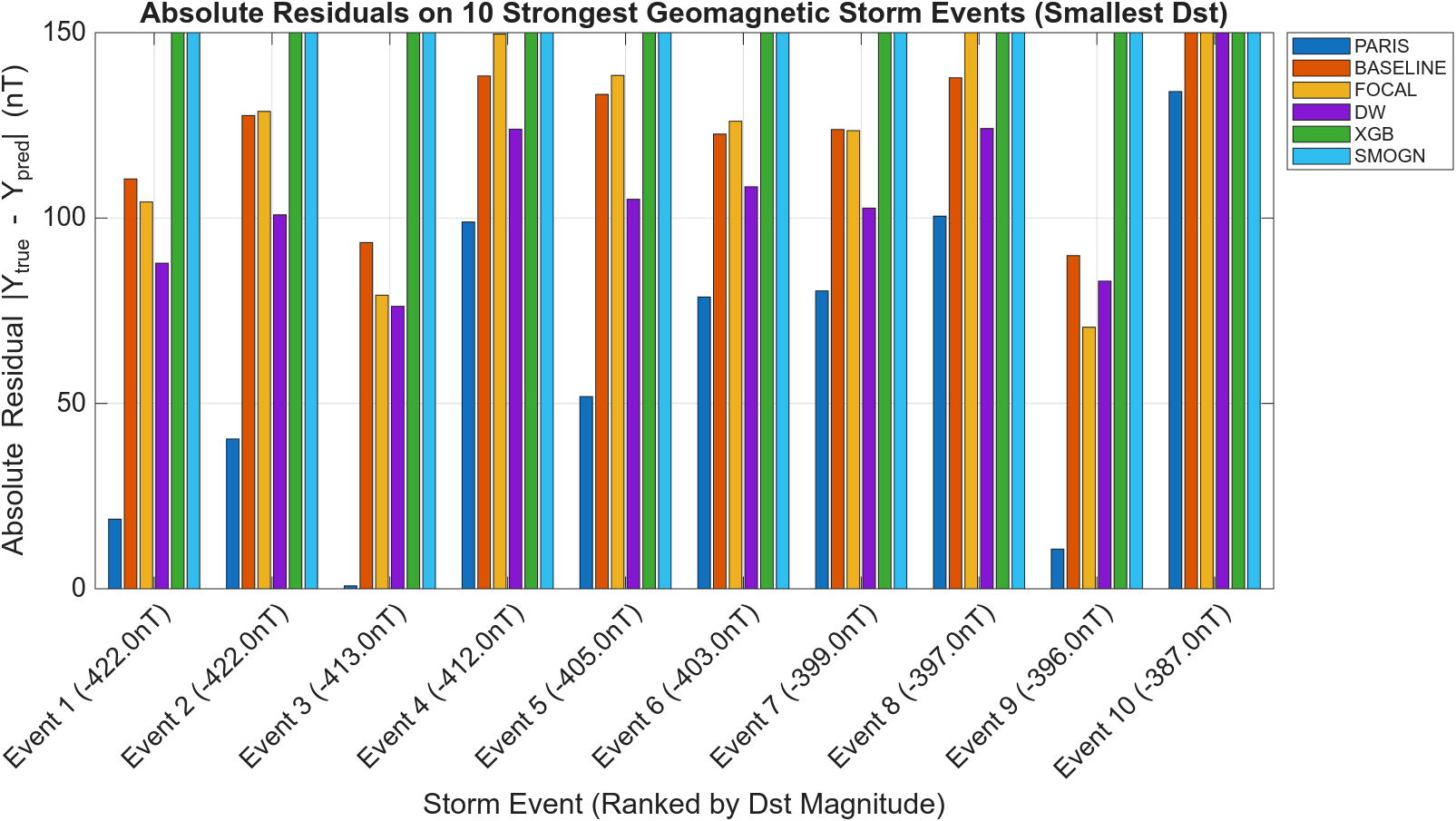}
    \caption{Absolute error for the 10 strongest storm values across the whole dataset. The vertical axis is linear, but cut off at 150nT (XGB and SMOGN exceed this value for all cases). For event 3 (-413.0 nT) PARIS' absolute residual is smaller than 1 nT.}
    \label{fig:strong_storms}
\end{figure}

Finally, we present in Figure \ref{fig:strong_storms} a bar chart that compares the absolute error for each model for the 10 strongest storm values across the whole dataset. Once again, PARIS outperforms all other model, yet using a fraction ($\sim 25\%$) of the original dataset size.

\section{Conclusion and Future Work}
The prevailing paradigm in deep learning research posits that model performance scales with the size of the training dataset. Under this view, providing sufficiently large amounts of data (as exemplified by the current emphasis on foundation models) will eventually lead to improved generalization. However, this strategy is inefficient in real-world settings where the primary interest lies in rare or extreme events (i.e., natural hazards, weather, climate, epidemiology, finance, etc.) In such scenarios, continuously adding high-frequency but uninformative samples can in fact \emph{degrade} performance on the low-frequency regimes that matter most.

In this work, we pursue the opposite philosophy: rather than enlarging the dataset, we iteratively \emph{prune} it, removing uninformative or performance-deteriorating samples while retaining a minimal but maximally informative subset for training. 

We introduced the \textbf{Pruning Algorithm via the Representer theorem for Imbalanced Scenarios (PARIS)}, grounded in the classical representer theorem. We derive a simple and computationally inexpensive estimator for the leave--one--out change in validation loss, $\Delta \mathcal{L}$, which quantifies the expected impact of pruning an individual training sample. This estimator yields a direct and efficient mechanism to identify and remove detrimental samples in batches, without requiring retraining after each removal.
By combining Cholesky rank--1 downdates with a structured update of the
representer influence matrix, our method avoids recomputing the dual coefficients $\alpha$ or
the influence matrix $S$ from scratch at every iteration.

Our empirical validation on the critical problem of Geomagnetic $D_{st}$ index forecasting demonstrates the effectiveness of PARIS. The method achieved the lowest overall RMSE, significantly outperforming state-of-the-art loss re-weighting and data augmentation baselines. Most importantly, PARIS excelled at predicting the severe, rare events, as evidenced by its superior performance across the conditional RMSE (cRMSE) curve, all while training on only 25\% of the original dataset. This confirms that the right subset of training data, identified structurally, is more valuable than artificially inflated or simply larger datasets.

The principled framework introduced in this work for estimating the change in validation loss $\mathcal{\Delta L}$ and for constructing the representer influence matrix $S$ opens several avenues for future research. First, analyzing the structure of $S$ may provide new insight into the physical mechanisms that determine why certain samples are more informative than others. In strongly non-linear systems, proximity in physical space does not necessarily correspond to proximity in the target space, and influence patterns encoded in $S$ could reveal these subtle relationships.

Moreover, we have not explored the regime in which the training set is sufficiently large that mini-batch training becomes unavoidable, which in turn requires redefining or approximating $S$ under stochastic feature updates. Understanding how PARIS behaves under large-scale or streaming-data conditions remains an important direction.

Additional opportunities for extending the PARIS framework include: (1) generalizing to multi-output regression or classification by adapting the residual-based pruning criterion to vector-valued losses; (2) incorporating influence information from multiple ``hard’’ validation points simultaneously, rather than relying solely on the single worst-case residual.

Overall, the interpretability and computational efficiency of PARIS suggest that influence-based pruning may become a broadly useful paradigm for training models in imbalanced, extreme-event, or data-scarce scientific regimes.

\newpage
\appendix
\section{Closed-form surrogate estimation of the ridge parameter \texorpdfstring{$\lambda$}{lambda}}
\label{app:lambda}

The PARIS algorithm requires the solution of the regularized normal equations
\[
(\Phi^\top \Phi + \lambda I_D) w = \Phi^\top (y - b),
\]
where $w$ and $b$ are the last-layer weights and bias of the trained neural network,
$\Phi$ is the feature matrix extracted from the current model, and $y$ are the training targets.
The regularization parameter $\lambda$ directly affects the conditioning of the primal system
and the numerical stability of the Cholesky downdate used throughout PARIS.  
Selecting $\lambda$ by cross-validation would require repeated retraining of the feature extractor,
which is infeasible inside an iterative pruning loop.  
Instead, we estimate an \emph{effective} ridge parameter using a lightweight closed-form surrogate.

Let \(w_{\mathrm{nn}}\in\mathbb{R}^D\) be the trained final-layer weight vector and
\(b_{\mathrm{nn}}\) its bias.  Define the centered targets
\[
Y_c = Y - b_{\mathrm{nn}}.
\]
Let
\[
A = \Phi^\top \Phi, 
\qquad 
b = \Phi^\top Y_c .
\]
If \(w_{\mathrm{NN}}\) were exactly the solution of the ridge system
\[
(A + \lambda I) w{\mathrm{NN}} = b,
\]
then rearranging yields
\[
\lambda w_{\mathrm{NN}} = b - A w_{\mathrm{NN}}.
\]
Projecting both sides onto \(w_{\mathrm{NN}}\) gives the scalar identity
\[
\lambda 
= \frac{w^\top_{\mathrm{NN}} (\,b - Aw_{\mathrm{NN}}\,)}{\|w_{\mathrm{NN}}\|_2^2}.
\]
We use this expression as a an estimator:
\[
\boxed{
\hat\lambda 
= 
\frac{
w_{\mathrm{NN}}^\top (\,b - A w_{\mathrm{NN}}\,)
}{
\| w_{\mathrm{NN}} \|_2^2
}}.
\]

If the denominator $\|w_{\mathrm{nn}}\|_2^2$ is too small, or if the resulting $\hat\lambda$
is non-positive or numerically unstable, we fall back to a conservative lower bound
$\lambda_{\min} = 10^{-5}$.  
This prevents ill-conditioned Cholesky factorizations and ensures robust behavior during pruning.

This estimator is extremely cheap to compute: it requires only a matrix--vector product
and solves no linear systems.  
Because $\Phi$ and $w_{\mathrm{nn}}$ are already available after training, the surrogate
introduces negligible overhead and avoids the need for cross-validation or iterative search.
Empirically, the resulting $\hat\lambda$ is stable and consistent across pruning iterations,
and it provides sufficient regularization for the Cholesky downdate process.

\paragraph{Acknowledgments.}
This work was partially supported by NASA under awards No. 80NSSC23M0192 and 80NSSC20K1580.

\paragraph{Disclosure of Interests.}
The author has no competing interests to declare that are relevant to the content of this article.


\end{document}